\documentclass[letterpaper]{article} 
\usepackage{aaai2026}  
\usepackage{times}  
\usepackage{helvet}  
\usepackage{courier}  
\usepackage[hyphens]{url}  
\usepackage{graphicx} 
\usepackage{natbib}  
\usepackage{caption} 
\usepackage{algorithm}
\usepackage{algorithmic}
\usepackage{multirow}
\usepackage[table]{xcolor}
\usepackage{booktabs}
\usepackage{amssymb}
\usepackage{subcaption}
\usepackage{amsmath}
\usepackage{newfloat}
\usepackage{listings}
\DeclareCaptionStyle{ruled}{labelfont=normalfont,labelsep=colon,strut=off} 
\floatstyle{ruled}
\newfloat{listing}{tb}{lst}{}
\floatname{listing}{Listing}
\title{A Dual-Phase Self-Evolution Framework for Large Language Models}
\author{
    Haoran Sun, 
    Zekun Zhang,
    Shaoning Zeng\\
}
\usepackage{bibentry}

\begin{document}
\maketitle

\begin{abstract}

The capabilities of Large Language Models (LLMs) are limited to some extent by pre-training, so some researchers optimize LLMs through post-training. Existing post-training strategies, such as memory-based retrieval or preference optimization, improve user alignment yet fail to enhance the model's domain cognition. To bridge this gap, we propose a novel Dual-Phase Self-Evolution (DPSE) framework that jointly optimizes user preference adaptation and domain-specific competence. DPSE introduces a Censor module to extract multi-dimensional interaction signals and estimate satisfaction scores, which guide structured data expansion via topic-aware and preference-driven strategies. These expanded datasets support a two-stage fine-tuning pipeline: supervised domain grounding followed by frequency-aware preference optimization. Experiments across general NLP benchmarks and long-term dialogue tasks demonstrate that DPSE consistently outperforms Supervised Fine-Tuning, Preference Optimization, and Memory-Augmented baselines. Ablation studies validate the contribution of each module. In this way, our framework provides an autonomous path toward continual self-evolution of LLMs.


\end{abstract}


\section{Introduction}
Large language models (LLMs), pre-trained on massive text corpora, have demonstrated remarkable general capabilities \cite{LLMsurvey2,LLMsurvey5}. However, their fixed parameters pose limitations in adapting to evolving user needs and domain-specific requirements \cite{LLMsurvey3,LLMsurvey6}. To address this, researchers have developed various post-training techniques aimed at enhancing LLM performance beyond pretraining \cite{LLMsurvey3}. Among them, preference optimization plays a pivotal role in aligning model outputs with human expectations \cite{INoT,datasetagent}. Early approaches employed reinforcement learning from human feedback (RLHF) \cite{RLHF}, where a learned reward model simulates user preferences to guide policy updates. More recent methods such as Direct Preference Optimization (DPO) \cite{DPO} and its variants bypass the complexity of RL by directly aligning the model’s output distribution with preference-labeled data \cite{evolving}. Since these methods heavily rely on large-scale human-annotated preferences, newer works have proposed iterative preference optimization, which can autonomously generate and refine training data during optimization, reducing the need for manual supervision \cite{UPO}. In parallel, other studies have integrated long-term memory mechanisms into LLM agent frameworks \cite{Amem,memorybank}. By incorporating user inputs with historical interaction traces into few-shot prompts, these approaches enhance the model's response quality and user alignment without modifying its underlying parameters \cite{memorysurvey,readagent}.

\begin{figure}[t]
    \centering
    \includegraphics[width=1\linewidth]{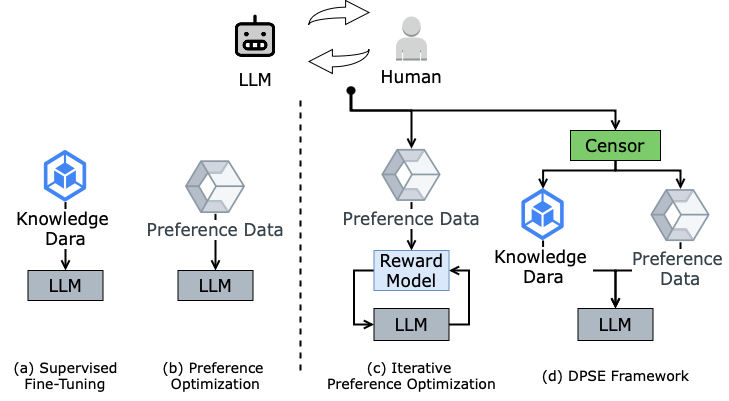}
    \caption{Overview of Dual-Phase Self-Evolution framework and comparison with other related work. }
    \label{fig:intro}
\end{figure}

However, these methods primarily focus on aligning LLM outputs with user preferences, while overlooking improvements to the models’ intrinsic cognitive capabilities \cite{,DPOsurvey}. As a result, they remain limited in achieving true self-evolution. In other words, such approaches emphasize user alignment—enhancing dialogue fluency and adaptability—yet fail to address the systematic advancement of the model’s core abilities in areas such as task completion and logical reasoning. Although Supervised Fine-Tuning (SFT) \cite{SFT1} has shown effectiveness in improving model performance on specific tasks, it heavily relies on large-scale, high-quality human-annotated datasets, incurring substantial cost and resource overhead \cite{SFT2}.

\begin{figure*}[t]
    \centering
    \includegraphics[width=1\linewidth]{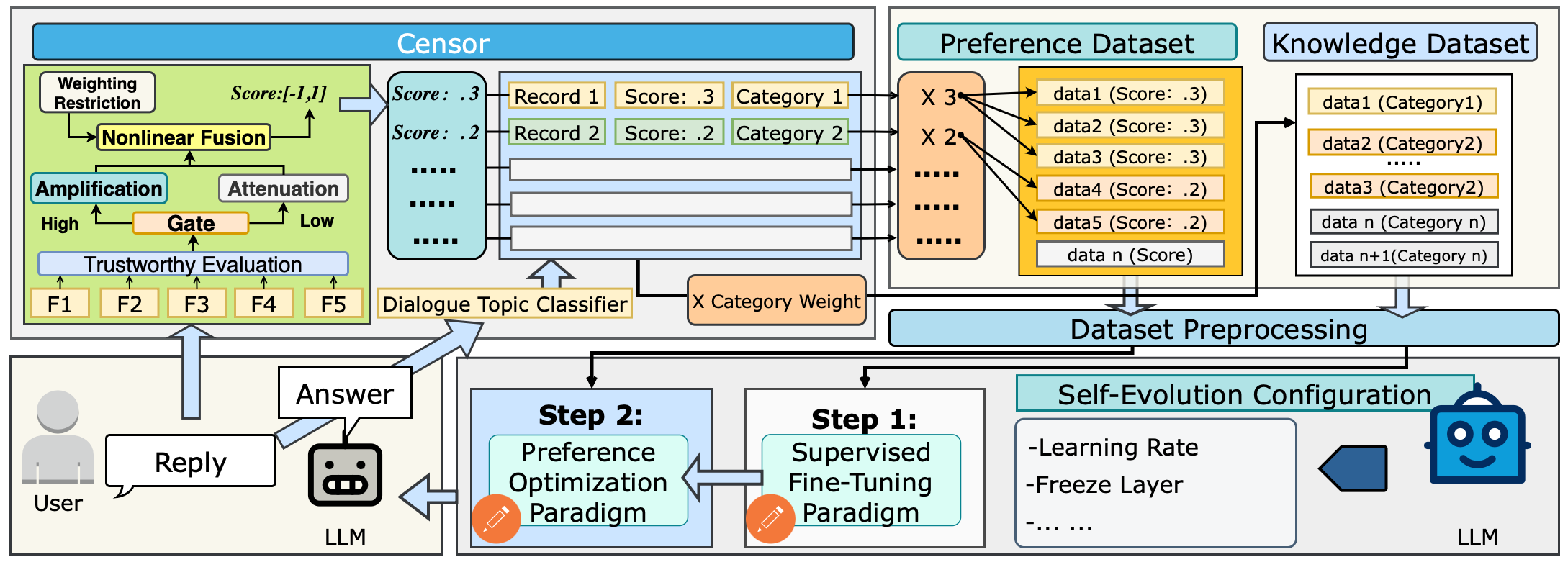}
    \caption{\textbf{Illustration of DPSE Framework. }Censor extracts multidimensional signals from user–model interactions, computes satisfaction scores, and performs topic classification to construct structured preference memory. Based on this memory, DPSE introduces two data expansion strategies, preference-driven expansion guided by satisfaction scores, and topic-aware expansion based on topic distribution. Subsequently, DPSE conducts two-stage training: supervised fine-tuning on domain-specific data, followed by preference optimization using satisfaction-labeled samples.}
    \label{fig:framework}
\end{figure*}

To address the limitations of fixed pre-trained weights in LLMs, we propose a Dual-Phase Self-Evolution (DPSE) framework that jointly enhances user preference alignment and domain-specific competence. DPSE features a signal-driven Censor module and a dual-phase fine-tuning pipeline, enabling autonomous evolution through structured data expansion. Extensive experiments on general NLP benchmarks and long-term dialogue tasks demonstrate that DPSE consistently outperforms strong baselines from SFT, PO, and Memory-Augmented frameworks. Our analysis further reveals the essential role of each component through detailed ablation studies.

\section{Related Work}

\textbf{Preference Optimization. }To align large language models (LLMs) with human preferences, preference optimization (PO) frameworks have been widely adopted \cite{DPOsurvey}. Early methods, such as reinforcement learning from human feedback (RLHF) \cite{RLHF}, trained reward models and used Proximal Policy Optimization (PPO) for alignment. Later approaches, including Direct Preference Optimization (DPO) \cite{DPO} and Rank from Human Feedback (Rrhf) \cite{rrhf}, directly optimized LLMs using human-labeled preference data. The PRO algorithm further extends this by incorporating preference ranking and multi-dimensional comparisons \cite{PRO}. However, collecting high-quality preference data remains costly and labor-intensive, motivating a shift toward automated and iterative optimization methods. 
Yet, these methods may introduce noisy preference pairs during iteration, hindering performance improvement. To mitigate this issue, the Uncertainty-enhanced Preference Optimization (UPO) framework \cite{UPO} filters reliable preference data using estimator models and MC Dropout techniques. Despite progress in enhancing user preferences in LLM outputs, these methods still have limitations in acquiring preference data and neglect the improvement of LLMs' own cognitive abilities.

\subsubsection{Supervised Fine-Tuning. }Supervised Fine-Tuning (SFT) is widely used to enhance large language models’ domain-specific expertise by training on annotated domain data \cite{SFT1}. It improves the model’s understanding of professional terminology and context, boosting accuracy and robustness in specialized tasks \cite{SFT2}. However, SFT requires extensive manual data collection and intervention during fine-tuning, leading to high resource costs. Thus, an automated framework for efficient data acquisition and fine-tuning is urgently needed to reduce these costs.

\subsubsection{Long-Term Memory. }Some researchers view conversational memory mechanisms as a form of self-evolution for large language models (LLMs) \cite{multidialogue}. Early methods concatenated full dialogue histories into prompts to preserve context, but were limited by the model’s context window and unsuitable for long-term interactions. To address this, more efficient memory systems have been developed. ReadAgent \cite{readagent} summarizes long texts for on-demand retrieval, improving context utilization. MemoryBank \cite{memorybank} uses vector-based similarity search to enhance storage and access, though scalability remains an issue. A-MEM \cite{Amem} constructs an evolving knowledge graph, improving organization at the cost of structural complexity. While differing in implementation, these methods all enhance user alignment via few-shot prompts that combine past interactions with current inputs \cite{LLMsurvey1}. However, they operate externally and do not improve the LLM’s internal cognition, which remains fixed and non-evolving.

\section{Methodology}

This section details the architecture and mechanisms of DPSE, including the Censor module for satisfaction estimation and topic classification, the dual-phase data augmentation strategies, and the two-phase fine-tuning process. Figure \ref{fig:framework} provides an overview of the framework. Additionally, DPSE mitigates hallucinations during topic classification and dataset expansion by combining prompt constraints, output validation, and uncertainty-based filtering. This multi-layered mechanism effectively excludes unreliable or inconsistent outputs, ensuring data quality and stabilizing subsequent model fine-tuning.

\subsection{Censor Module}
Censor module identifies and filters responses that genuinely satisfy the user during discussions on a given topic, storing them in the memory module to support subsequent self-evolution.

\subsubsection{Signal Extraction and Preprocessing. }To support interpretable satisfaction estimation, the Censor module extracts five representative signals from user–model interactions, capturing both behavioral and semantic cues. These signals are as follows.

\begin{itemize}
    \item \textbf{Explicit Feedback. }A binary indicator (1 for praise, 0 otherwise), extracted via rule-based heuristics from logs or comments. It represents direct user preference and is used as a reliable anchor signal with a lower-bound weight in the fusion stage.

    \item \textbf{Dwell Time. }The duration the user spends reading the model response, discretized into three levels (short = 0, medium = 1, long = 2). To account for the non-linear relationship between time and attention, we apply a U-shaped transformation:

    \begin{equation}
    f(dwell) = -0.5 \cdot (dwell - 1)^2 + 0.5
    \end{equation}

    This function performs a non-linear transformation in the user's dwell time in the answer area, with a value range of $[0, 2]$, presenting a symmetric U-shaped structure, whose maximum value is 0.5 and occurs when $dwell$ = 1. This mapping function serves as an indirect quantitative signal for "degree of attention" in satisfaction estimation.

    \item \textbf{Coherence. }Measured as the cosine similarity between the embeddings of the user query and the model’s response. High coherence reflects better fluency, relevance, and semantic consistency.
    
    \item \textbf{Similarity. }Captures semantic redundancy by comparing the current response with previous ones in the conversation. Excessively high similarity is penalized, especially under negative sentiment, as it often suggests repetition or lack of novelty.
    
    \item \textbf{Sentiment. }The predicted emotional polarity of the model’s response (positive, neutral, or negative), obtained from a sentiment classifier. It not only serves as an independent signal but also modulates the influence of other signals, particularly the similarity score.
    
\end{itemize}

\subsubsection{Dynamic Gating and Credibility Evaluation.}After preprocessing, the five normalized signals are passed through a credibility-aware gated fusion module that dynamically controls their influence in satisfaction estimation. This mechanism integrates two neural sub-networks to capture both signal salience and trustworthiness.

The Gate Network, implemented as an MLP with Sigmoid activation, produces a gating vector $G \in [0, 1]^5$, which softly controls the activation of each signal. The Credibility Network, also an MLP with Softmax activation, outputs a normalized weight vector $C \in \Delta^5$, representing the relative reliability of each signal.
The two vectors jointly modulate the input signals $S \in \mathbb{R}^5$, yielding the gated and trust-weighted signals,
\begin{equation}
\text{GatedSignals} = G \odot S \odot e^{C},
\end{equation}
where $\odot$ denotes element-wise multiplication and $\exp(C)$ emphasizes highly credible signals. This fusion step allows for fine-grained, dynamic control over signal contribution, combining salience and credibility in a unified formulation.


\subsubsection{Incorporating Physical Constraints. }To ensure that satisfaction modeling remains interpretable and aligned with human intuition, we introduce a set of physically inspired constraints on fusion weights. These constraints regulate the influence of each signal in a data-adaptive yet human-consistent manner.

\noindent(1) Fixed budget for redundancy signal
Among the five signals, semantic similarity plays a distinct role as a penalizer for redundancy. To reflect its limited interpretive power compared to other features, we cap its weight at 10\% of the total, assigning the remaining 90\% to the other four:
\begin{equation}
\begin{aligned}
\tilde{w}_i = 0.9 \cdot \frac{w_i}{\sum_{j \neq \text{sim}} w_j}, &\quad i \in \{\text{fb}, \text{dwell}, \text{coh}, \text{sent}\} \\
&\tilde{w}_{\text{sim}} = 0.1
\end{aligned}
\end{equation}
(2) Sentiment-modulated similarity penalty
When the user expresses negative sentiment, high content similarity is more likely to indicate redundancy or user dissatisfaction. Thus, we dynamically modulate the similarity weight by the negative sentiment score:
\begin{equation}
\tilde{w}_{\text{sim}} \leftarrow \tilde{w}_{\text{sim}} \cdot \sigma(-\beta \cdot s_{\text{sent}}),
\end{equation}

where $\sigma$ is the Sigmoid function and $\beta > 0$ controls sensitivity.

\noindent (3) Minimum influence of explicit feedback
To preserve the effect of direct user feedback regardless of signal noise, we enforce a minimum threshold $\tau > 0$ for the explicit feedback weight:
\begin{equation}
\tilde{w}_{\text{fb}} \leftarrow \max(\tilde{w}_{\text{fb}}, \tau)
\end{equation}
(4) Weight normalization
After constraint application, the adjusted weights are rescaled to ensure a valid convex combination:
\begin{equation}
\hat{w}_i =  \begin{cases} \frac{\tilde{w}_i}{\sum_j \tilde{w}_j}, & \text{if } \sum_j \tilde{w}_j > 1 \\ \tilde{w}_i, & \text{otherwise} \end{cases}
\end{equation}
(5) Final scoring
The final satisfaction score is computed as a weighted sum over the transformed input signals $\tilde{s}_i$, using the constrained and normalized weights $\hat{w}_i$:
\begin{equation}
\text{Score}_{\text{sat}} = \sum_i \hat{w}_i \cdot \tilde{s}_i
\end{equation}

\noindent This physically-constrained fusion strategy improves interpretability, avoids overfitting to spurious features, and encodes intuitive behavioral assumptions (e.g., “explicit praise matters”, “redundancy under negativity is bad”), thereby enhancing both robustness and generalization.

\subsubsection{Nonlinear Fusion and Satisfaction Scoring. }Signals processed through dynamic gating, confidence enhancement, and constraint-based weighting are passed to a nonlinear fusion layer (fusion\_net)—a lightweight neural network that performs complex transformations and outputs a single satisfaction score. This score is normalized to the [-1, 1] range via a Tanh activation: positive values indicate satisfaction, negative values indicate dissatisfaction, and zero denotes neutrality. The nonlinear fusion captures intricate interactions among signals, enabling comprehensive satisfaction assessment.


\subsubsection{Memory Classification. }To support structured domain-specific data construction, the Censor module incorporates a lightweight topic classifier. It takes the concatenated user query and model response as input, performs semantic and pragmatic analysis, and assigns a topic label from a predefined set (e.g., Medical, Sports) using a compact LLM backbone \cite{classification}. The prompt is: "Please classify the following input into one of the predefined categories. Do not add explanations or extra text." Each conversation is then represented as a triple: content, satisfaction score, and domain category.

\subsection{Dataset Construction and Preprocessing}
Although the Censor module filters high-quality, preference-aligned samples, the limited scale of real-world user–LLM interactions remains a bottleneck for fine-tuning. To overcome this, DPSE introduces a dual expansion mechanism that automatically constructs two datasets: one for domain-specific supervised fine-tuning and another for preference-based optimization. When the memory pool reaches a predefined threshold N, the system retrieves stored interactions and triggers both expansions based on satisfaction scores and topic distributions.


\noindent \textbf{(1) Expansion based on satisfaction scores. }To generate high-quality data aligned with user preferences, we adopt a score-proportional linear expansion strategy. Each stored sample is duplicated in proportion to its satisfaction score—for instance, a score of 0.4 results in 4 copies, while 0.2 results in 2. This biases training toward highly satisfying interactions and suppresses noise from low-confidence examples.

Additionally, new samples are generated using the base LLM guided by preference-aware prompts, which preserve the original semantics while varying tone, structure, and length to simulate user-oriented diversity. A representative instruction is:
“Given a user–assistant interaction, generate N responses that maintain the original intent and fluency while aligning with user satisfaction." This approach emphasizes output fluency, tone preference, and satisfaction density, forming the preference-optimized dataset used in DPSE’s second training phase.

\noindent \textbf{(2) Expansion based on topic categories. }To preserve domain diversity and prevent overfitting to high-scoring examples, we implement a category ratio–aware expansion strategy. The system computes long-term topic distributions from stored samples and expands underrepresented topics accordingly to ensure balanced coverage.

For each topic, representative samples are selected and prompts are constructed to generate semantically diverse yet stylistically consistent assistant responses within the same domain. A representative instruction is: “Given a conversation and its topic label, generate alternative responses that remain within the topic but vary in reasoning path or linguistic style.” The resulting domain-balanced dataset enhances knowledge grounding and generalization, and is used for the supervised fine-tuning phase of DPSE.

\subsection{Dual-Phase Self-Evolution}

DPSE employs a two-stage fine-tuning pipeline to simultaneously enhance the model’s domain-specific reasoning and user preference alignment. 


\noindent \textbf{(1) Supervised Fine-Tuning. }The first stage targets domain cognition through supervised training on the topic-expanded dataset. This step enables the model to learn correct task formats, reasoning patterns, and domain-specific knowledge under strong supervision signals. Following best practices from InstructGPT \cite{instructionGPT} and LLaMA-2 \cite{llama}, we perform instruction tuning by minimizing the cross-entropy loss over ground-truth outputs. This stage stabilizes the model’s behavior and reduces the risk of factual drift during subsequent preference optimization. Without domain grounding, directly optimizing for user satisfaction may lead to stylistically pleasing but factually incorrect outputs, particularly in knowledge-intensive fields such as medicine or law.

\noindent \textbf{(2) Preference Optimization. }Once the model acquires task-relevant knowledge through supervised fine-tuning, we apply Direct Preference Optimization (DPO) to align the model’s behavior with user preferences. DPO offers a stable, gradient-based alternative to RLHF and has been shown to outperform reward modeling in aligning generation with human preferences. Unlike standard DPO, which treats all training pairs equally, we incorporate satisfaction-aware frequency weighting to emphasize stronger user-preference signals. Specifically, each retained dialogue sample is assigned a satisfaction score $s_i \in [0, 1]$ by the Censor module. This satisfaction score is scaled by a constant $K$ (e.g., 10) and floored to obtain the expansion frequency $f_i$, which determines how many times sample $i$ is duplicated for training. Higher satisfaction leads to more frequent inclusion, e.g., a score of 0.4 leads to four copies. These frequencies are used in two ways:

(1) Pairwise sampling: Preference pairs are constructed from generated variants by comparing high-score vs. low-score responses. A sample with more duplications is more likely to appear in pairwise training, encouraging the model to learn from frequently satisfying behavior. (2) Gradient weighting: Frequencies directly modulate the DPO loss to prioritize preference-dense regions. The resulting weighted loss function is:

\begin{equation}
\mathcal{L}_{\text{wDPO}}(\theta) = - \sum_{i=1}^{N} \frac{w_i}{\sum_{j=1}^{N} w_j} \cdot \log \sigma \left( s_\theta(x_i, y_i^+) - s_\theta(x_i, y_i^-) \right)
\end{equation}

Among them, $\theta$ represents the model parameters, $(x_i, y_i^+, y_i^-)$ is the preference pair of the $i$-th training sample, including the input $x_i$, the user's preferred answer $y_i^+$, and the less preferred answer $y_i^-$; $s_\theta(x, y)$ represents the scoring function of the model for generating output $y$ from input $x$, usually taken as $\log p_\theta(y|x)$; $\sigma(\cdot)$ is the Sigmoid function, used to convert the score difference into a ranking probability; $w_i$ is the sample sampling weight obtained by the satisfaction score expansion mechanism, reflecting its preference frequency in the training data; $N$ represents the total number of preference pairs in the training dataset. The corresponding parameter layer is:

\begin{align}
\nabla_\theta \mathcal{L}_{\text{wDPO}}(\theta)
&= - \sum_{i=1}^{N} \frac{w_i}{\sum_{j=1}^{N} w_j} \cdot \nabla_\theta \log \sigma \big( \nonumber \\
&\quad s_\theta(x_i, y_i^+) - s_\theta(x_i, y_i^-) \big)
\end{align}

This strategy effectively introduces implicit supervision of "preference intensity" while keeping the learning mechanism of DPO itself unchanged. Through dataset construction and improved DPO, we have both intuitively shaped the preference density distribution and endowed the model with different response sensitivities to different preference levels during the training process.

\subsubsection{Training Strategy. }We propose a general fine-tuning paradigm that abstracts low-level configurations while allowing direct access to generated datasets. The framework enables large models to auto-adjust key training parameters—such as learning rate, batch size, and frozen layers—based on resource constraints (e.g., GPU memory, device count) and dataset traits. 




\section{Experiment}

\subsection{Setup}

\noindent \textbf{Baselines. }Following the practice in previous works \cite{UPO,Amem}, to thoroughly evaluate DPSE, we compare it with representative methods from three categories: supervised fine-tuning (SFT), preference optimization (PO), and memory-based approaches. PO baselines include DPO and UPO; memory-based methods include ReadAgent (RA), MemoryBank (MB), and A-MEM (AM). Both SFT and PO optimize model weights post-training—SFT improves domain knowledge, while PO aligns with user preferences. DPSE unifies these through dual-phase self-evolution. We conduct unified comparisons against SFT/PO to assess joint optimization, and separate comparisons with memory-based methods, which rely on external retrieval rather than parameter updates, to contrast endogenous (parameter tuning) and exogenous (memory retrieval) evolution.


\subsubsection{Dataset and Evaluation Metrics. }Following prior work, we evaluate the DPSE framework against SFT and preference optimization (PO) baselines on general NLP tasks, and against memory-focused baselines on long-term dialogue tasks. For general NLP, we use AlpacaEval 2.0 \cite{alpacaEval} and MT-Bench \cite{MT-Bench}, and for long-term dialogue, the LoCoMo dataset \cite{LoCoMoDataset}. AlpacaEval 2.0 contains 805 instruction-following questions from five sources and uses GPT-4-Turbo as an automatic judge to compare model responses against references. We report both raw win rate (WR) and length-controlled (LC) win rate. MT-Bench evaluates LLMs across eight fundamental capabilities (e.g., writing, reasoning, coding, STEM, humanities) on a 0–10 scale, and we report absolute scores. LoCoMo assesses long-term memory in multi-session dialogue. We select two representative subsets: (1) Single-hop (SH.) questions (2705 pairs) and (2) Multi-hop (MH.) questions (1104 pairs). Performance is measured using F1 score (capturing accuracy and completeness) and BLEU-1 score (measuring lexical precision).

\subsubsection{Implementation Details. }For general NLP tasks, we use Zephyr-7B (zephyr-7b-sft-full) \cite{Zephyr}, instruction-tuned on UltraChat200K, as the backbone. Following prior work, UltraFeedback \cite{ultrafeedback} and UltraChat200K \cite{ultrachat200k} serve as DPSE’s input for self-evolution, which is disabled during benchmark evaluation. Baselines include Zephyr-7B fine-tuned by SFT and DPO, plus UPO-Merge (single-round) and UPO (multi-round) variants. DPSE autonomously sets its SFT and preference optimization parameters. For memory-based comparisons, we follow prior settings using Qwen2-1.5B/7B \cite{qwen2} core models. Unlike non-finetuned baselines, DPSE evolves online during deployment by collecting real-time data from the LoCoMo dataset, which targets long-term, multi-turn dialogues. We test three self-evolution trigger thresholds—500 ($DPSE^1$), 1000 ($DPSE^2$), and 2000 ($DPSE^3$)—to analyze performance and identify the optimal setting. All baselines strictly follow their original training protocols. DPSE’s augmentation and prompting adhere to this work’s specifications. To ensure fairness, all models use only the question and instruction fields without extra prompt engineering. Experiments are conducted locally via Ollama.


\begin{table}[h]
    \centering
  
  \begin{tabular}{c|c|c}
    \toprule  
   \multirow{1}*{ \textbf{Methods}}  
    &\multicolumn{1}{c|}{ \textbf{AlpacaEval 2.0}}

    &\multirow{1}*{\textbf{MT-Bench}}

 \\
    
  \toprule
  
  Zephyr-7B-SFT& 5.84 &6.18  \\

  Zephyr-7B-DPO&9.12&6.79\\
  Zephyr-7B-UPO &13.04&7.02\\
Zephyr-7B-UPO-Merge&12.04&6.85\\
\cellcolor{gray!25}Zephyr-7B-\textbf{DPSE} 1 & \cellcolor{gray!25}12.98& \cellcolor{gray!25}7.69*\\
\cellcolor{gray!25}Zephyr-7B-\textbf{DPSE} 2 & \cellcolor{gray!25}\textbf{14.26*}&\cellcolor{gray!25}\textbf{8.46*}\\
\cellcolor{gray!25}Zephyr-7B-\textbf{DPSE} 3 & \cellcolor{gray!25}13.37*&\cellcolor{gray!25}8.14*\\

    \toprule
\end{tabular}

 \caption{Experimental results compared with SFT and PO baselines, which are obtained from GPT-4 automatic evaluation on AlpacaEval 2.0 (LC-weighted win rate \% relative to the GPT-4 reference) and MT-Bench (absolute scores). The best performance in each category is highlighted in bold. }
\label{tab:conpared_to_SFTPO}
\end{table}

  

    
    
  


\begin{table}[h]
    \centering
  
  \begin{tabular}{ccc|cccc}
    \toprule  
   \multicolumn{3}{c|}{ \multirow{2}*{\textbf{Models}}}  
    
    &\multicolumn{2}{c}{ \textbf{SH.}}&\multicolumn{2}{c}{ \textbf{MH.}}
    
\\
\cline{4-5}
\cline{6-7}
    &&&\textbf{F1} &\textbf{B1} &\textbf{F1} &\textbf{B1} \\
    
       \toprule
  \multirow{10}*{\rotatebox{90}{\textbf{Qwen2}}} &\multirow{5}*{\rotatebox{90}{\textbf{1.5b}}}  
  &RA. &6.25 &4.23 &4.67 &3.12  \\
  &&MB. &10.87 &8.21 &9.29 &7.77  \\
  &&AM. &18.01 &11.67 &24.87 &19.25 \\
  
&&\cellcolor{gray!25}\textbf{DPSE} 1 & \cellcolor{gray!25}21.25*& \cellcolor{gray!25}13.21*& \cellcolor{gray!25}29.38*&\cellcolor{gray!25}23.16*\\
&&\cellcolor{gray!25}\textbf{DPSE} 2 & \cellcolor{gray!25}\textbf{23.44*}& \cellcolor{gray!25}\textbf{15.61*}& \cellcolor{gray!25}\textbf{34.12*}&\cellcolor{gray!25}\textbf{29.89*}\\
&&\cellcolor{gray!25}\textbf{DPSE} 3 & \cellcolor{gray!25} 21.36*& \cellcolor{gray!25}13.65* & \cellcolor{gray!25} 30.12*&\cellcolor{gray!25}24.57* \\

  \cline{2-7}
  &\multirow{5}*{\rotatebox{90}{\textbf{7b}}}&RA. &3.42 &2.48 &3.14 &3.25  \\
  &&MB. &3.55 &3.12 &9.36 &8.97  \\
  &&AM. &12.29 &9.24 &26.13 &24.12   \\
 &&\cellcolor{gray!25}\textbf{DPSE} 1 & \cellcolor{gray!25}19.37*& \cellcolor{gray!25}12.23*& \cellcolor{gray!25}31.25*&\cellcolor{gray!25}24.36\\
 &&\cellcolor{gray!25}\textbf{DPSE} 2 & \cellcolor{gray!25}\textbf{22.65*}& \cellcolor{gray!25}\textbf{16.24*}& \cellcolor{gray!25}\textbf{33.52*}&\cellcolor{gray!25}\textbf{27.46*}\\
 &&\cellcolor{gray!25}\textbf{DPSE} 3 & \cellcolor{gray!25}21.68*& \cellcolor{gray!25}14.26*& \cellcolor{gray!25}31.55*&\cellcolor{gray!25}24.34\\
\toprule

\end{tabular}

 \caption{\textbf{Experimental results compared with Memory baselines. }We evaluate multiple methods using F1 and BLEU-1 (B1) scores (in \%). The best performance in each category is highlighted in bold. }
\label{tab:conpared_to_memory}
\end{table}

\subsection{Main Result and Analysis}

Each result represents the average over three independent runs with different random seeds. We conducted paired t-tests between DPSE and the best-performing baseline. Results marked with $\textbf{*}$ indicate statistically significant improvements (p $<$ 0.05).

\subsubsection{Comparison to SFT and PO baselines. }As shown in Table \ref{tab:conpared_to_SFTPO}, the experimental results show that DPSE, under three different self-evolution trigger thresholds, consistently outperforms the SFT and PO baselines, with win rates improving from 13.04\% to 14.26\%. On the MT-Bench benchmark, where GPT-4 is used to annotate average scores across eight evaluation dimensions, DPSE again demonstrates superior performance, with scores increasing from 7.02\% to 8.46\%. The comparison results with SF and PO baselines demonstrate the superiority of the dual-phase self-evolution. Furthermore, during the self-evolution phase, DPSE integrates both UltraFeedback and UltraChat200K as input, and subsequent evaluations on distinct datasets confirm its strong transferability and robustness. We analyzed that the reason DPSE outperforms the SF and PO baselines is that DPSE takes into account both user preferences and domain professional knowledge, integrating the advantages of both, thus enabling a more comprehensive optimization of LLM capabilities.

\subsubsection{Comparison to Memory baselines. }As shown in Table \ref{tab:conpared_to_memory}, we conducted comparative experiments with memory baselines on the LoCoMo dataset, using F1 and BLUE-1 to evaluate the performance of the method. It can be seen that DPSE outperforms the baselines on both F1 and BLUE-1, indicating that DPSE not only improves the accuracy of LLM answers but also enhances the quality, demonstrating the effectiveness of the dual-phase evolution. In addition, we conducted experiments using 1.5b and 7b LLMs respectively, and the results show that DPSE can effectively optimize LLMs of different scales, eliminating the potential interference that LLM scale may have on the experiments. We attribute DPSE’s superior performance over memory-based baselines to two main factors. First, memory operates as an external framework and does not enhance the LLM’s internal capabilities, thus being constrained by the model's original limitations. Second, memory mechanisms rely solely on prompt-level integration of historical interactions, which may improve alignment with user preferences but offer limited enhancement to domain-specific expertise.

\begin{figure}[t]
    \centering
    \includegraphics[width=1\linewidth]{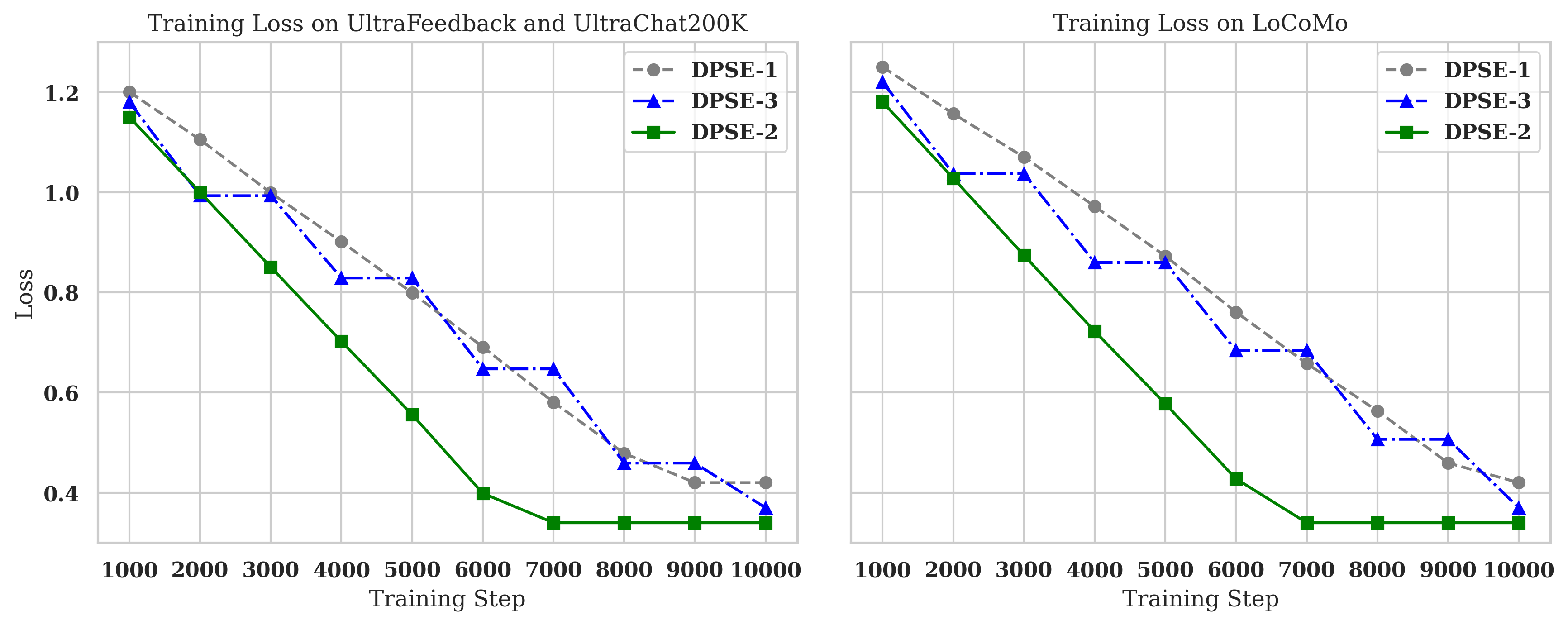}
    \caption{The curve of training loss on UltraFeedback, UltraChat200K and LoCoMo at each Trigger Threshold. }
    \label{fig:curve}
\end{figure}

\section{Further Analysis}

\begin{figure*}[t]
    \centering
    \includegraphics[width=1\linewidth]{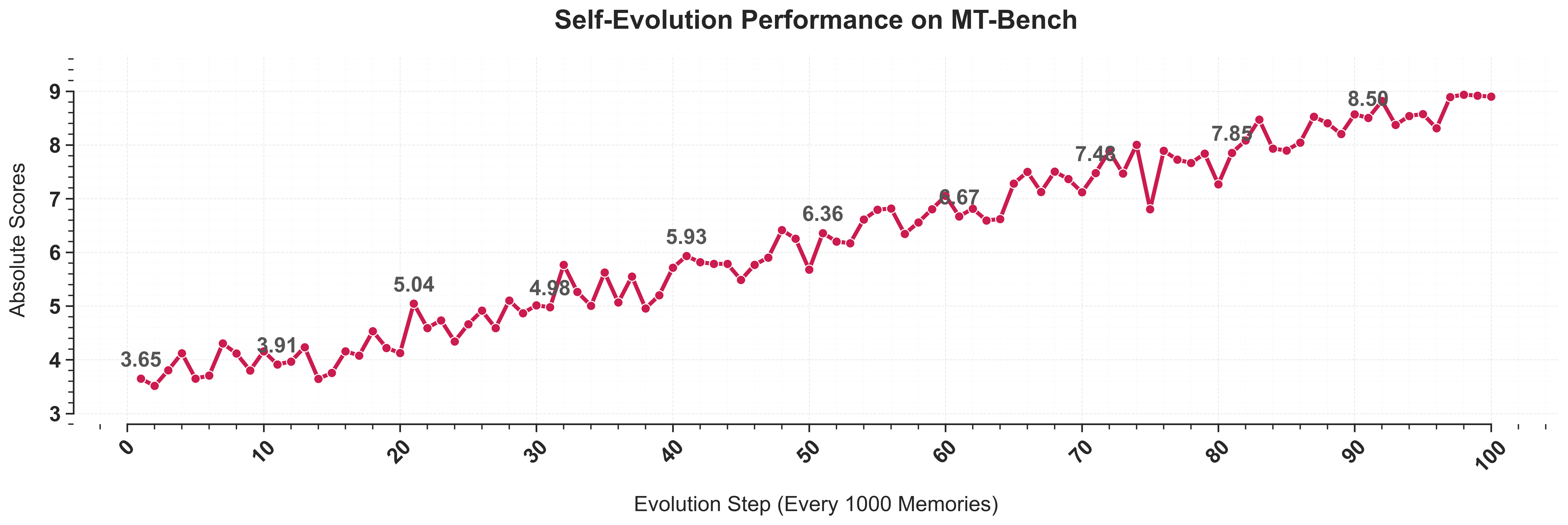}
    \caption{We input the WildChat dataset into DPSE to simulate long-term usage by real users and validate it on the MT-Bench dataset (absolute score).}
    \label{fig:long_term}
\end{figure*}

\begin{table*}[h]
    \centering
  
  \begin{tabular}{p{11cm}|ccc}
    \toprule  
   \multirow{1}*{ \textbf{Feedback}}  
   &\multicolumn{1}{c}{ \textbf{Type}}
   
    &\multicolumn{1}{c}{ \textbf{Signal}}

    &\multirow{1}*{\textbf{Score}}

 \\
    
       \toprule
  “Very clear!” no follow-up; answer is relevant and concise.  &Positive &[1,1,0.92,-0.2,1]& 0.74 \\
  \toprule
  No comment; long dwell; response is somewhat relevant but redundant.& Moderate&[0,2,0.75,-0.4,0]& 0.52 \\
  \toprule
  
  “This doesn’t answer my question.”; short dwell; response is off-topic.& Negative &[0,0,0.38,-0.6,-1]& 0.18\\
  \toprule

  Asked “What about semi-supervised methods?” &Follow-up&[0,2,0.81,-0.5,-0.2]&0.38\\

    \toprule
\end{tabular}
 \caption{Analysis of Censor's effectiveness to judge satisfaction score of user feedback. }
\label{tab:censor_analysis}
\end{table*}

\begin{table}[h]
    \centering
 \begin{minipage}[c]{0.50\textwidth}
    \centering
  
  \begin{tabular}{c|cc}
    \toprule  
   \multirow{1}*{ \textbf{Models}}

   &\multicolumn{1}{c}{ \textbf{AlpacaEval 2.0}}
  
    &\multicolumn{1}{c}{ \textbf{MT-Bench}}  
\\ 
       \toprule
   w/o. CS. &7.23& 6.24 \\
  w/o. DC.&6.23 & 7.17\\
  (SFT) w/o. SE. &9.34 & 6.78\\ 
  (PO) w/o. SE. &9.24&6.13\\
DPSE 2& \textbf{14.26}&\textbf{8.46}\\
\toprule
\end{tabular}
  \begin{tabular}{c|cccc}
    \toprule  
   \multirow{2}*{ \textbf{Models}}  
&\multicolumn{2}{c}{ \textbf{SH.}}&\multicolumn{2}{c}{ \textbf{MH.}}
\\
\cline{2-3}
\cline{4-5}
    &\textbf{F1} &\textbf{BLEU-1} &\textbf{F1} &\textbf{BLEU-1} \\
       \toprule
   w/o. CS. &14.23&9.78  &28.35&22.14 \\
  w/o. DC.& 14.78&10.24&29.36&21.47 \\
  (SFT) w/o. SE. &17.34 &12.14&30.28& 24.16\\ 
  (PO) w/o. SE. &16.22&11.34&31.10&23.68\\
DPSE 2& \textbf{23.44}&\textbf{15.61}&\textbf{34.12}&\textbf{29.89}\\

\toprule
\end{tabular}

\end{minipage}
       
 \caption{\textbf{Ablation Study. }Specially, we choose the results on LoCoMo dataset using Qwen2-1.5B. CS. represents Censor Module, DC. means Dataset Construction Module and SE. is Self-Evolution Module. }
\label{tab:ablation}
\end{table}

\noindent \textbf{Optimal Trigger Threshold Analysis. }As shown in Tables \ref{tab:conpared_to_SFTPO} and \ref{tab:conpared_to_memory}, experiments on three benchmarks indicate that a trigger threshold of 1,000 achieves the best self-evolution performance. Figure \ref{fig:curve} shows that $DPSE^2$ demonstrates the most stable convergence, with steadily decreasing loss and minimal fluctuations. Although $DPSE^3$ achieves a larger overall loss reduction, its longer update intervals cause slower early-stage decline and slightly less stability than $DPSE^2$. $DPSE^1$, with frequent updates but low per-update gain, shows limited overall improvement and higher final loss. A threshold of 2,000 leads to large but sparse updates that may miss key signals at low data density, while 500 causes frequent but noisy updates with low gains. Thus, 1,000 balances update frequency and effectiveness, ensuring continuous training and sufficient improvement.

\subsubsection{Analysis of Censor's Effectiveness. }Real-world human behavior is complex, and no dataset fully simulates human reaction on LLM responses. To evaluate Censor’s effectiveness, we designed questions and collected four feedback types from real users, then scored them using Censor. Table \ref{tab:censor_analysis} shows an example: the question “What’s the difference between supervised and unsupervised learning?” and the LLM’s answer. Censor successfully extracted five feedback signals, assigning scores of 0.74 for positive and 0.18 for negative cases. For neutral and follow-up feedback without explicit signals, Censor integrated dwell time, semantic overlap, and sentiment to assign reasonable intermediate scores (0.52 and 0.38). These results demonstrate Censor’s ability to detect both explicit and implicit feedback, including follow-up questions and semantic repetition, reflecting strong satisfaction recognition.

\subsubsection{Analysis of the Self-Evolution in long-term real-world using. }WildChat \cite{wildchat} collected 1 million conversations between human users and ChatGPT by offering free access to GPT-3.5 and GPT-4. These authentic user interactions realistically reflect the distribution of user queries and simulate long-term real-world usage. We randomly sampled 100,000 entries from WildChat and continuously fed them into DPSE, simulating long-term user interaction, with the self-evolution trigger threshold set to 1,000. After each self-evolution, we validated on the MT-Bench dataset, using absolute scores to assess effectiveness. As shown, the MT-Bench absolute score rose from 3.65 to 8.97, and the LLM performance in DPSE consistently improved over time, demonstrating the effectiveness of self-evolution in long-term use.

\subsubsection{Ablation Study. }To validate the effectiveness of DPSE’s three core modules, we performed ablation studies by removing each module individually and measuring performance impact. Without the Censor module, raw data bypasses quality filtering before evolution. Without Dataset Construction, filtered data is used directly without augmentation. Without Self-Evolution, training relies solely on standard supervised fine-tuning (SFT) or preference optimization (PO) with default hyperparameters. As Table \ref{tab:ablation} shows, performance declines consistently when any module is removed, confirming each module’s essential role. Notably, removing Censor or Dataset Construction causes greater degradation than removing Self-Evolution, highlighting the critical importance of data quality for effective fine-tuning.



\section{Conclusion}

In this paper, we propose a Dual-Phase Self-Evolution (DPSE) framework that jointly enhances user preference alignment and domain-specific competence. DPSE features a signal-driven Censor module and a dual-phase fine-tuning pipeline, enabling autonomous evolution through structured data expansion. To evaluate the effectiveness of DPSE, we designed extensive experiments and compared it with SFT, PO, and memory methods, and the results demonstrated that DPSE can achieve high-quality self-evolution of LLMs.


\bibliography{main}

\end{document}